\documentclass[runningheads]{llncs}
\usepackage{graphicx}
\usepackage{multirow}       
\usepackage[acronym]{glossaries} 
\usepackage{mathtools}          
\usepackage{algorithm}
\usepackage[noend]{algpseudocode}
\usepackage{enumitem}

\usepackage[utf8]{inputenc} 
\usepackage[T1]{fontenc}    
\usepackage{hyperref}       
\usepackage{url}            
\usepackage{booktabs}       
\usepackage{amsfonts}       
\usepackage{nicefrac}       
\usepackage{microtype}      
\usepackage{xcolor}         
\usepackage{cite}
\algnewcommand{\LeftComment}[1]{\(\hfill \triangleright\) \texttt{#1}}
\algnewcommand{\BoldComment}[1]{\(\hfill \triangleright\) \underline{\emph{#1}}}
\algnewcommand{\CenterComment}[1]{\texttt{//#1}}

\newcommand{\cmark}{\checkmark}
\newcommand{\xmark}{$\times$}

\newacronym{fl}{FL}{Federated Learning}
\newacronym{sgd}{SGD}{Stochastic Gradient Descent}
\newacronym{bn}{BN}{Batch Normalization}
\newacronym{erm}{ERM}{Empirical Risk Minimization}
\newacronym{fa}{FA}{Federated Adaptation}
\newacronym{da}{DA}{Domain Adaptation}
\newacronym{dnn}{DNN}{Deep Neural Network}
\newacronym{dg}{DG}{Domain Generalization}
\newacronym{pda}{PDA}{Post-Deployment Adaptation}

\algtext*{EndFunction}

\begin{document}
\title{Post-Deployment Adaptation with Access to Source Data via Federated Learning and Source-Target Remote Gradient Alignment}
\titlerunning{Post-Deployment Adaptation via FL and StarAlign}
%
\author{Felix Wagner\inst{1} \and Zeju Li\inst{2} \and Pramit Saha\inst{1} \and Konstantinos Kamnitsas\inst{1}}
\authorrunning{F. Wagner et al.}
%
\institute{Department of Engineering Science, University of Oxford, UK \and
Nuffield Department of Clinical Neurosciences, University of Oxford, UK \\ 
\email{felix.wagner@eng.ox.ac.uk}}
\maketitle              
\begin{abstract}
Deployment of Deep Neural Networks in medical imaging is hindered by distribution shift between training data and data processed after deployment, causing performance degradation. Post-Deployment Adaptation (PDA) addresses this by tailoring a pre-trained, deployed model to the \emph{target} data distribution using limited labelled or entirely unlabelled target data, while assuming no access to \emph{source} training data as they cannot be deployed with the model due to privacy concerns and their large size. This makes reliable adaptation challenging due to limited learning signal.
This paper challenges this assumption and introduces \textbf{FedPDA}, a novel adaptation framework that brings the utility of learning from remote data from Federated Learning into PDA. FedPDA enables a deployed model to obtain information from source data via remote gradient exchange, while aiming to optimize the model specifically for the target domain. Tailored for FedPDA, we introduce a novel optimization method \texttt{StarAlign} (\textbf{S}ource-\textbf{Ta}rget \textbf{R}emote Gradient \textbf{Align}ment) that aligns gradients between source-target domain pairs by maximizing their inner product, to facilitate learning a target-specific model.
We demonstrate the method's effectiveness using multi-center databases for the tasks of cancer metastases detection and skin lesion classification, where our method compares favourably to previous work. Code is available at: \url{https://github.com/FelixWag/StarAlign}

\end{abstract}
\section{Introduction}\label{intro}

Effectiveness of \Glspl{dnn} relies on the assumption that training (source) and testing (target) data are drawn from the same distribution (domain). When \Glspl{dnn} are applied to target data from a distribution (target domain) that differs from the training distribution (source domain), i.e. there is distribution shift, \Glspl{dnn}' performance degrades \cite{kamnitsas2017unsupervised,geirhos2018degrationstudy3}. For instance, such shift can occur, when a pre-trained \Gls{dnn} model is deployed to a medical institution with data acquired from a different scanner or patient population than the training data. This hinders reliable deployment of \glspl{dnn} in clinical workflows.\\
\textbf{Related Work:} Variety of approaches have been investigated to alleviate this issue.
\Gls{dg} approaches assume access to data from multiple source domains during training. They aim to learn representations that are invariant to distribution shift, enabling better generalization to any unseen domain \cite{muandet2013domain,li2018learning,dou2019domain}. Due to the great heterogeneity in medical imaging, achieving universal generalization may be too optimistic.
Instead, \Gls{da} methods \cite{ben2010theory,ganin2016domain,kamnitsas2017unsupervised} aim to learn a model that performs well on a specific target domain. They assume that (commonly unlabelled) target data  and labelled source data is collected in advance and centrally aggregated to train a model \emph{from scratch}. Privacy concerns in healthcare limit the scalability of these approaches.

\Gls{fl} \cite{McMahan2017fedavg} enables training a single model on multiple decentralised databases. The original Federated Averaging algorithm (\texttt{FedAvg})\cite{McMahan2017fedavg} and its extensions target scenarios where different "nodes" in the federated \emph{consortium} have data from different source domains \cite{li2020fedProx,karimireddy20aSCAFFOLD,Zhao2018fedNonIID,li2021fedbn}. They train a single model to generalize well across unseen domains. 
In contrast, approaches for \emph{Personalization} in \Gls{fl} \cite{arivazhagan2019federatedPersonalization,liu2021feddg,li2021fedbn,Jiang2023IOPFl} learn multiple models, one for each source domain but do not support adaptation to unseen target domains.

Federated Domain Adaptation (FDA) methods \cite{peng2019federated,li2020multi,feng2021KD3A} enable \Gls{da} in a federated setting. These methods assume abundant unlabelled data on the target domain and perform distribution matching to enforce domain invariance. They train a model from scratch for each target which limits their practicality, while the enforced invariance can lead to loss of target-specific discriminative features.

\begin{figure}[t]
    \includegraphics[width=\textwidth]{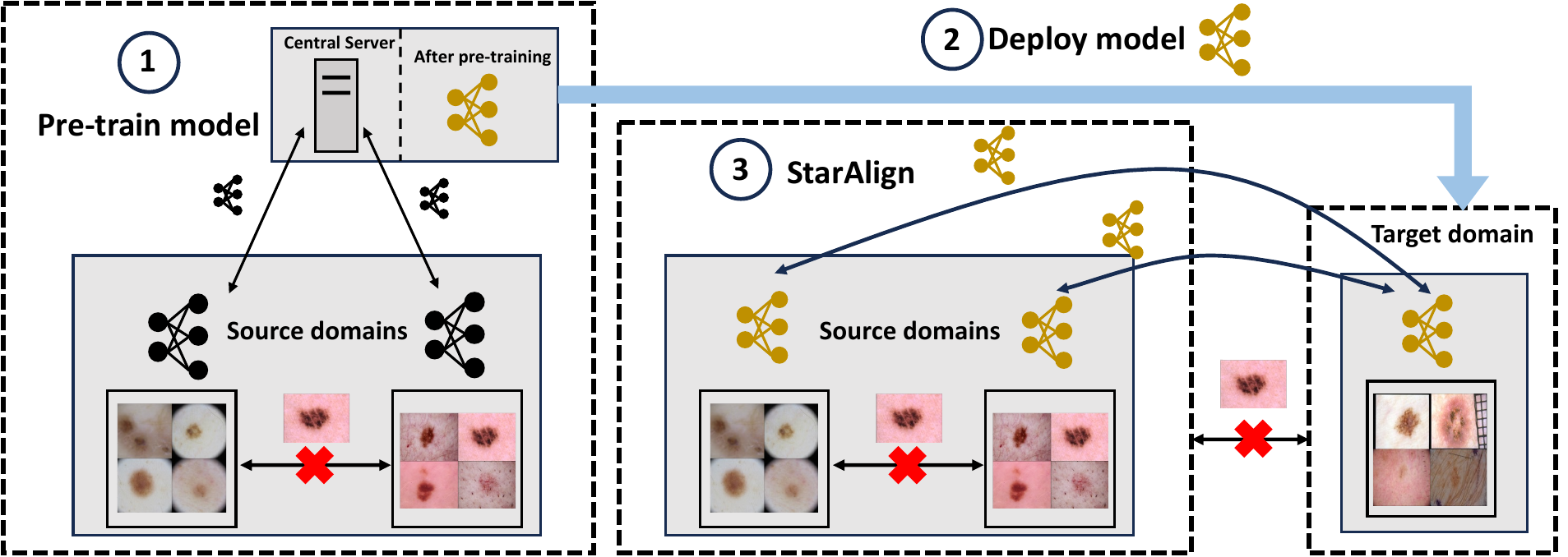}
    \caption{The \textbf{FedPDA} framework consists of three steps: (1) The model is trained on source domains (here via FL); (2) The model is deployed to the target domain; (3) Adaptation is done via gradient alignment between all source-target domain pairs with \texttt{StarAlign} without cross-domain data exchange.}
    \label{fig:StarAlignFramework}
\end{figure}

\Gls{pda} methods (or Test-Time Adaptation or Source-free DA) seek to enable a pre-trained/deployed model to optimize itself for the specific target domain by learning only from limited labelled or unlabelled target data processed after deployment \cite{wang2020tent,karani2021test,valvano2021stop,chidlovskii2016sourcefreedomain,chen2021source,bateson2022source}. These methods assume that deployed models have no access to the source data due to privacy, licensing, or data volume constraints. In practice, however, adaptation using solely limited labelled or unlabelled data can lead to overfitting or unreliable results.\\
\textbf{Contribution:} This work presents a novel \Gls{pda} framework to \emph{optimize a pre-trained, deployed model} for a \emph{specific target domain} of deployment, assuming \emph{limited labelled data at the deployment node}, while \emph{remotely obtaining information from source data} to facilitate target-specific adaptation.
\begin{itemize}
\item The framework enables a deployed DNN to obtain information from source data without data exchange, using remote gradient exchange. This overcomes PDA's restriction of unavailable access to source data. This combines \Gls{fl} with \Gls{pda} into a new framework, \textbf{FedPDA} (Fig.~\ref{fig:StarAlignFramework}). Unlike \Gls{fl}, FedPDA optimizes a target-specific model instead of a model that generalises to any distribution (\Gls{fl}) or source distributions (personalized \Gls{fl}). While \Gls{da} requires central aggregation of data, FedPDA does not transfer data. FedPDA also differs from FDA by adapting a \emph{pre-trained deployed} model at the user's endpoint, rather than training a model \emph{pre-deployment from scratch} with abundant unlabelled target data and ML developer's oversight. This addresses the practical need for reliable adaptation with limited data and technical challenge of optimizing from a source-specific initial optimum to a target-specific optimum.

\item Tailored specifically for this setting, we introduce the \texttt{StarAlign} optimization algorithm (\textbf{S}ource-\textbf{Ta}rget \textbf{R}emote Gradient \textbf{Align}ment) for decentralised multi-domain training. It extracts gradients from source data and aligns them with target data gradients by maximising the gradient inner product, to regularize adaptation of a target-specific model.
\end{itemize}
We evaluate the method through extensive experiments on two medical datasets: cancer metastases classification in histology images using Camelyon17 with 5 target domains \cite{bandi2018detection}, and a very challenging setting for skin lesion classification using a collection of 4 databases (HAM \cite{tschandl2018ham10000}, BCN \cite{combalia2019bcn20000}, MSK \cite{codella2018skin}, D7P \cite{kawahara2018seven}).
In numerous settings, our method achieves favourable results compared to state-of-the-art PDA, (personalized) \Gls{fl}, FDA and gradient alignment methods.

\section{Background}\label{sec:Background}

\textbf{Problem setting:}
We consider the general case with a set of \(S\) source domains \(\{ \mathcal{D}_1, \dots, \mathcal{D}_S \}\) and one target domain \(\mathcal{D}_{T}\), where \(T\!=\!S\!+\!1\). Each domain's dataset \(\mathcal{D}_k \coloneqq \{ (x_i^k, y_i^k) \}_{i=1}^{n_k} \) is drawn from a data distribution \(p_k\), where $x_i$ is an image and $y_i$ the corresponding label. We assume there are little available labels in the target domain, $|\mathcal{D}_{T}|\ll |\mathcal{D}_{k}|, \forall k \in [1,S]$.

In a setting with multiple source domains, optimizing a DNN's parameters $\theta$ is commonly done with \Gls{erm}, by minimizing a cost $R_K$ for each domain $\mathcal{D}_k$, which is the expectation of a loss $\mathcal{L}$ over $\mathcal{D}_k$:
\begin{equation} \label{eq:fedGlobalObjective}
\min_{\theta}R_{ERM}(\theta) \coloneqq \frac{1}{S}\sum_{k=1}^{S}R_{k}(\theta) =  \frac{1}{S}\sum_{k=1}^{S}\mathbb{E}_{(x,y)\sim\mathcal{D}_k} [\mathcal{L}(\theta;x,y)].
\end{equation}
This can be performed via centralised training (e.g. \Gls{dg} methods) or approximated via FL when \(\{ \mathcal{D}_1, \dots, \mathcal{D}_S \}\) are distributed and without data sharing. In our experiments we focus on the latter challenging case. We assume model $\theta$ is pre-trained and `deployed' to a target domain where limited labelled $D_T$ is available.
Due to domain shift $p_T\neq p_k, \forall k \in [1,S]$, we assume $\theta$ may not generalize to the target domain. Next, we describe our algorithm for adapting $\theta$ to \(p_T\).

\section{Method}
\label{method}

Using only limited labelled target data \(\mathcal{D}_{T}\) for PDA is likely to give unreliable results. Therefore, our FedPDA framework, enables a deployed model to obtain information from source data using remote gradient exchange via FL (Fig.~\ref{fig:StarAlignFramework}). To achieve this, our \texttt{StarAlign} algorithm for FedPDA aligns gradients of all domains with target domain gradients by maximising their inner product.
This ensures that source-derived gradients promote learning features that are relevant to the target domain.

\subsection{Theoretical Derivation of Source-Target Alignment}
\label{subsec:centralStar}

First, we derive our algorithm in the \emph{theoretical} setting where source and target data \(\{ \mathcal{D}_1, \dots, \mathcal{D}_S, \mathcal{D}_T\}\) are available for centralised training. The subsequent section presents our \emph{distributed algorithm} for remote domains, developed to approximate the optimization (Eq.~\ref{eq:fStarAlign}) in the theoretical setting.
We form pairs between each domain and target $\mathcal{D}_T$ to extract target-relevant information from each source domain. For the $k$-th pair we minimise the combined cost \( R_{kT}(\theta)\!=\!R_{k}(\theta)\!+\!R_{T}(\theta)\).
To align gradients \(G_k = \mathbb{E}_{\mathcal{D}_k} [\nabla_{\theta} \mathcal{L}_k(\theta; x,y)]\) of the $k$-th domain
and the target domain \(G_T = \mathbb{E}_{\mathcal{D}_T} [\nabla_{\theta} \mathcal{L}_T(\theta; x,y)]\), we maximize their inner product $G_T\! \cdot\! G_k$. Hence over all pairs, we minimize the total cost:
\begin{equation}\label{eq:fStarAlign}
    R^{\texttt{StarAlign}}_{total} = \frac{1}{S\!+\!1} \sum_{k=1}^{S+1} R^{\texttt{Align}}_{kT}(\theta) = \frac{1}{S\!+\!1} \sum_{k=1}^{S+1} \left( R_{kT}(\theta) - \delta G_T \cdot G_k \right),
\end{equation}

with \(\delta\) a hyperparameter. Using dot product's distributive property we get: 
\begin{equation}\label{eq:AllSourceDomainPairs}
    R^{\texttt{StarAlign}}_{total} = R_{T}(\theta) + \frac{1}{S\!+\!1} \sum_{k=1}^{S+1}R_{k}(\theta)- \delta G_T \cdot \sum_{k=1}^{S+1} \frac{G_k}{S\!+\!1}
\end{equation}

This allows us to interpret the effect of optimizing Eq.~\ref{eq:fStarAlign}. We minimize the target domain's cost, regularized by the average cost over all domains, which mitigates overfitting the limited data $\mathcal{D}_T$. The third term forces the average gradient over all domain data, $\sum_{k=1}^{S+1} \frac{G_k}{S\!+\!1}$, to align with the gradients from the target domain $G_T$. We clarify that domain pairs summed in Eq.~\ref{eq:fStarAlign} include the pair target-to-target with $R_{TT}$ and $G_T\cdot G_T$ terms. This ensures that the average gradient of all domains $\sum_{k=1}^{S+1} \frac{G_k}{S\!+\!1}$ in Eq.~\ref{eq:AllSourceDomainPairs} always has a component along the direction of $G_T$, avoiding zero or very low dot products, for instance due to almost perpendicular gradients in high dimensional spaces, leading to smoother optimization trajectory. We found this effective in preliminary experiments.

Directly optimising dot-products in Eq.~\ref{eq:fStarAlign} is computationally expensive as it requires second-order derivatives. Instead, minimizing Eq.~\ref{eq:fStarAlign} is approximated via Alg. \ref{alg:CentralStargAlign} and first-order derivatives in \emph{L4-L9}. This is based on the result in \cite{shi2022fish} that the update step \(\beta (\hat{\theta} - \theta)\) (\emph{L9}) approximates optimisation of Eq.~\ref{eq:fStarAlign} for a pair of domains. \(\beta\) is a scaling hyperparameter. This was used in \cite{shi2022fish} for \emph{centralised} multi-source \Gls{dg}, aligning gradients of \emph{source-source} pairs to learn a general model (\emph{no target}). Here, we minimize Eq.~\ref{eq:fStarAlign} to align \emph{source-target} pairs of gradients and learn a \emph{target-specific} model.

\begin{algorithm}[t]
\caption{Centralised gradient alignment for source-target domain pairs}\label{alg:CentralStargAlign}
\small
\begin{algorithmic}[1]
\State $\theta_T$: Model pre-trained on source data \(\{\mathcal{D}_{1},\dots,\mathcal{D}_{S}\}\), deployed to Target
\For{\(j = 1,2 \dots \)}
\For{\(\mathcal{D}_k \in \{\mathcal{D}_1, \dots, \mathcal{D}_S, \mathcal{D}_T\}\)}
    \State \(\hat{\theta}_k \gets \theta_T\); \(\theta_k \gets \theta_T\) \quad\quad\quad \LeftComment{Create copies of deployed model $\theta_T$}
    \State \( d_T \sim \mathcal{D}_T \) \quad\quad\quad\quad\quad\quad\quad \LeftComment{Sample batch from domain \(T\)}
    \State \( \hat{\theta}_k \!\gets \! \hat{\theta}_k - \alpha \nabla_{\hat{\theta}_k} \mathcal{L}(\hat{\theta}_k;d_T) \) \LeftComment{Compute gradient of \(d_T\) and update model}
    \State \( d_k \sim \mathcal{D}_k \) \quad\quad\quad\quad\quad\quad\quad\ \LeftComment{Sample batch from domain \(k\)}
    \State \( \hat{\theta}_k \!\gets \hat{\theta}_k - \alpha \nabla_{\hat{\theta}_k} \mathcal{L}(\hat{\theta}_k;d_k)\) \LeftComment{Compute gradient of \(d_k\) and update model}
    \State \( \theta_k \gets \theta_k + \beta(\hat{\theta}_k-\theta_k) \) \quad\ \ \LeftComment{1st order approximation update}
\EndFor
\State \( \theta_T \gets \frac{1}{S+1}\sum_{k=1}^{S+1}\theta_k\) \quad\quad \LeftComment{Update model with avg. over domains}
\EndFor
\end{algorithmic}
\end{algorithm}

\subsection{Source-Target Remote Gradient Alignment for FedPDA}

We assume a model pre-trained on source data $\{\mathcal{D}_1,\dots,\mathcal{D}_S\}$ is deployed on a computational node with access to target data $D_T$ but not source data.
Below, we describe the case when each source dataset $\{\mathcal{D}_1,\dots,\mathcal{D}_S\}$ is held in a separate compute node (e.g. federation).
We now derive the distributed \texttt{StarAlign} Algorithm \ref{alg:DistrStarAlign}, approximating minimization of Eq.~\ref{eq:fStarAlign} without data exchange between nodes.

The aim is to approximate execution of Alg.~\ref{alg:CentralStargAlign} on the target node. Lack of access to source data, however, prevents calculation of source gradients in \emph{L8} of Alg.~\ref{alg:CentralStargAlign}. Instead, we approximate the gradient of each source domain $\mathcal{D}_s$ separately, by computing the average gradient direction \(\overline{g}_s\) on the specific source node. For this, we perform \(\tau\) local optimisation steps on source node $s$ and average their gradients, obtaining \(\overline{g}_s\). By transferring \(\overline{g}_s\) from each source node to the target node, we can approximate the updates in \emph{L8} of Alg.~\ref{alg:CentralStargAlign} on the target node.

\begin{algorithm}[ht]
    \caption{Distributed \texttt{StarAlign} algorithm for FedPDA}\label{alg:DistrStarAlign}
\begin{algorithmic}[1]
\State $\theta_T$: Model pre-trained on source data \(\{\mathcal{D}_{1},\dots,\mathcal{D}_{S}\}\), deployed to Target node
\For{\(j = 1\) to \(E\)} \quad\quad\quad\quad \LeftComment{where $E$ total communication rounds}
    \For{\(\mathcal{D}_s \in \{\mathcal{D}_{1}, \dots, \mathcal{D}_S \} \) \textbf{in parallel} } \quad \BoldComment{Performed at each Source node}
        \State \(\theta_s \gets \theta_T\)  \quad\quad\quad\quad\quad\quad\quad \LeftComment{Obtain latest model from Target node}
        \For{\(i = 1\) to \(\tau\)}   \quad\quad\quad \LeftComment{Local iterations}
            \State \( d_s \sim \mathcal{D}_s \) \quad\quad\quad\quad\quad\quad\quad \LeftComment{Sample batch}
            \State \( g_s^i \gets \nabla_{\theta_s} \mathcal{L}(\theta_s;d_s)\) \quad\quad\quad  \LeftComment{Compute gradient}
            \State \( \theta_s \gets \theta_s - \alpha g_s^i \) \quad\quad\quad\quad\quad \LeftComment{Update model}
        \EndFor
        \State \(\overline{g}_s=\frac{1}{\tau} \sum_{i=1}^\tau g_s^i \) \quad \BoldComment{Compute avg. gradient and \textbf{SEND} to Target node}
    \EndFor

    \For{ \( k \in \{1 ,\dots ,S, T\} \) }  \BoldComment{Code below is performed at Target node}
        \State \(\theta_k \gets \theta_T\); \(\hat{\theta}_k \gets \theta_T\) 
        \For{\(i = 1\) to \(\tau\)}  \quad\quad\quad\quad\quad\quad \LeftComment{Local iterations}
            \State \( d_T \sim \mathcal{D}_T \) \quad\quad\quad\quad\quad\quad\quad\quad\quad \LeftComment{Sample batch}
            \State \( \hat{\theta}_k \gets \hat{\theta}_k - \alpha \nabla_{\hat{\theta}_k} \mathcal{L}(\hat{\theta}_k;d_T) \) \quad \LeftComment{Compute gradient on \(d_T\) and update}
            \State \textbf{if} \(k \neq T\) \textbf{then} \( \hat{\theta}_k \gets \hat{\theta}_k - \alpha \overline{g}_k \) \LeftComment{Update with average gradient}
            \State \textbf{else} Repeat L13-14: Resample \( d_{T}^{\prime}\! \sim\! \mathcal{D}_T \), update \( \hat{\theta}_k \! \gets\! \hat{\theta}_k \!-\! \alpha \nabla_{\hat{\theta}_k} \mathcal{L}(\hat{\theta}_k;d_{T}^{\prime}) \)
        \EndFor
        \State \( \theta_k \gets \theta_k - \beta(\hat{\theta}_k - \theta_k)  \) \quad\quad\quad\quad \LeftComment{1st order approximation update}
    \EndFor
    \State \(\theta_T = \frac{1}{S+1} \sum_{k=1}^{S+1} \theta_k\) \BoldComment{Update \(\theta_T\) and \textbf{SEND} to Source nodes}
\EndFor
\end{algorithmic}
\end{algorithm}

Alg.~\ref{alg:DistrStarAlign} presents the distributed \texttt{StarAlign} method that aligns gradients between all source-target domain pairs. First, target and source node models get initialised with the pre-trained model \( \theta_T \) (\emph{L1}). Each source node $s$ computes its average gradient direction \(\overline{g}_s\) and communicates it to the target node (\emph{L3-L9}).
After receiving the average gradient directions of source domains, the target node performs the interleaving approximated updates (\textit{L14-15}) for source and target domain gradients for each domain pair for \( \tau \) steps. \emph{L16} implements the case of the target-target pair (Sec.~\ref{subsec:centralStar}) with a second actual update rather than approximation. Finally, \textit{L17} performs the first-order approximation update. Note, that we perform this update step after \(\tau\) steps, which we found empirically to give better results than performing it after each interleaving update.
This process is repeated for \(E\) communication rounds.

\section{Experiments}
\textbf{Cancer detection:} We use Camelyon17 \cite{bandi2018detection} dataset. The goal is to predict whether a histology image contains cancer (binary classification). The data was collected from 5 different hospitals, which we treat as 5 domains.\\
\textbf{Skin lesion diagnosis:} We train a model to classify 7 skin lesion types in dermatoscopic images. To create a multi-domain benchmark, we downloaded four public datasets acquired at 4 medical centers: HAM \cite{tschandl2018ham10000}, BCN \cite{combalia2019bcn20000}, MSK \cite{codella2018skin} and D7P \cite{kawahara2018seven}. Due to missing classes in MSK and D7P, we combine them in one that contains all classes. Therefore we experiment with 3 domains.

\begin{table}[t]
\centering
\caption{Test accuracy (\%) on tumour detection (3 seeds averaged).}
\label{tab:resultsCamelyon}
\resizebox{\textwidth}{!}{%
\begin{tabular}{@{}llcccccccc@{}}
\toprule
\multicolumn{2}{l}{Method} & \multicolumn{2}{c}{$D_T$} & \multicolumn{5}{l}{Target domain} &  \\ \midrule
Pre-Train & Adapt & Lab. & Unlab. & Hospital 1 & Hospital 2 & Hospital 3 & Hospital 4 & Hospital 5 & Average \\ \midrule
FedAvg & None  & \xmark & \xmark & 88.0 & 80.5 & 76.2 & 85.4 & 75.2 & 81.1 \\
None & From scratch & \cmark & \xmark & 92.7 & 88.4 & 92.9 & 91.3 & 94.7 & 92.0 \\
FedBN & None (avg. $\mathcal{D}_S$ BN stats) & \xmark & \xmark & 50.3 & 53.8 & 70.7 & 50.0 & 74.3 & 59.8 \\
FedBN & None ($\mathcal{D}_T$ BN stats) & \xmark & \cmark & 89.4 & 84.5 & 89.6 & 91.4 & 84.8 & 87.9 \\
FedBN & Fine-tuning & \cmark & \xmark & 93.1 & 88.2 & 93.7 & 95.8 & \textbf{97.2} & 93.6 \\
FedBN & Supervised TENT & \cmark & \xmark & 92.8 & 86.8 & 91.6 & 94.1 & 93.1 & 91.7 \\
FedBN & FedPDA: FedBN & \cmark & \xmark & 91.7 & 75.7 & 94.7 & 83.3 & 55.2 & 80.1 \\
FedBN & FedPDA: PCGrad & \cmark & \xmark & 93.8 & 90.0 & 94.9 & 95.1 & 95.0 & 93.8 \\
FedBN & FedPDA: StarAlign & \cmark & \xmark & \textbf{95.0} & \textbf{91.4} & \textbf{96.8} & \textbf{96.8} & 94.5 & \textbf{94.9} \\ \midrule
None & FDA-KD3A & \xmark & \cmark & 94.3 & \textbf{91.4} & 90.4 & 93.6 & 94.3 & 92.8 \\ \bottomrule
\end{tabular}%
}
\end{table}

\subsection{Experimental setup}
\textbf{Setup:} For all experiments we use DenseNet-121 \cite{huang2017densely}. We perform `leave-one-domain-out' evaluations, iterating over each domain. One domain is held-out as target and the rest are used as source domains. Each domain dataset is divided into train, validation and test sets (60,20,20\% respectively). The training sets are used for pre-training and in FedPDA for source domains. As we investigate how to adapt with limited labelled data in the target domain, we only use 1.8\% of a target domain's dataset as labelled for PDA on Camelyon17, and 6\% for skin-lesion. The process is then repeated for other target domains held-out.\\
\textbf{Metrics:} Each setting is repeated for 3 seeds. From each experiment we select 5 model snapshots with best validation performance and report their average performance on the test set (averaging $15$ models per setting). The two classes in Camelyon17 are balanced, therefore we report \textit{Accuracy} (Tab.~\ref{tab:resultsCamelyon}). Skin-lesion datasets have high class-imbalance and hence \emph{Accuracy} is inappropriate because methods can increase it by collapsing and predicting solely the majority classes. Instead, we report \emph{Weighted Accuracy} (Tab.~\ref{tab:resultsSkinLesion}) defined as: \( \sum_{c=1}^{C} \frac{1}{C} \text{acc}_c \), where \(C\) the number of classes and \(\text{acc}_c\) represents accuracy for class $c$.\\
\textbf{Methods:}
The `Pre-Train' column in Table \ref{tab:resultsCamelyon} and \ref{tab:resultsSkinLesion} indicates the pre-deployment training method, employing \texttt{FedAvg} \cite{McMahan2017fedavg} or \texttt{FedBN} \cite{li2021fedbn} with $\tau=100$ local iterations per communication round.
\texttt{FedBN} pre-training is used thereafter as it outperformed \texttt{FedAvg} when BN statistics were adapted (\textbf{$D_T$ BN stats} below).
Column `Adapt' indicates the post-deployment adaptation method. `Lab' and `Unlab' indicate whether labelled or unlabelled target data are used for adaptation. We compare \texttt{StarAlign} with: no adaptation (\textbf{None}), training the model just on target data (\textbf{from scratch}), \texttt{FedBN} without adaptation using average \Gls{bn} statistics from source domains (\textbf{avg. $D_S$ BN stats}) and when estimating \Gls{bn} stats on unlabelled $D_T$ data (\textbf{$D_T$ BN stats}) \cite{li2021fedbn}, and \textbf{fine-tuning} the model using only target labelled data. We also compare with the \Gls{pda} method \textbf{TENT} \cite{wang2020tent}, adapted to use labelled data (supervised) via cross entropy instead of unsupervised entropy for fair comparison, and the state-of-the-art FDA method \textbf{KD3A} \cite{feng2021KD3A}
(using whole training set as unlabelled).
We also attempt to perform FedPDA by simply integrating the target domain node into the federated system and resuming \texttt{FedBN} starting from the pre-trained model (\textbf{FedPDA: FedBN}), and with another gradient alignment method, a variant of \textbf{PCGrad} \cite{yu2020gradient}, which we made applicable for FedPDA. \texttt{PCGrad} projects gradients that point away from each other onto the normal plane of each other. \textbf{StarAlign} uses $\tau\!=\!100$, and $\beta\!=\!0.01$ or $\beta\!=\!0.2$ for Camelyon17 and skin lesion tasks respectively, configured on validation set.

\begin{table}[t]
\centering
\caption{Test weighted accuracy (\%) on skin lesion diagnosis (3 seeds averaged).}
\label{tab:resultsSkinLesion}
\scriptsize
\resizebox{0.85\textwidth}{!}{
\begin{tabular}{@{}llcccccc@{}}
\toprule
\multicolumn{2}{l}{Methods} & \multicolumn{2}{c}{$D_T$} & \multicolumn{3}{l}{Target domain} &  \\ \midrule
Pre-Train & Adapt & Lab. & Unlab. & BCN & HAM & MSK \& D7P & Average \\ \midrule
FedAvg & None & \xmark & \xmark & 26.0 & 37.6 & 25.2 & 29.6 \\
None & From scratch & \cmark & \xmark & 34.8 & 40.0 & 24.5 & 33.1 \\
FedBN & None (avg. $\mathcal{D}_S$ BN stats) & \xmark & \xmark & 26.6 & 34.8 & 24.3 & 28.6 \\
FedBN & None ($\mathcal{D}_T$ BN stats) & \xmark & \cmark & 35.4 & 39.7 & 29.5 & 34.9 \\
FedBN & Fine-tuning & \cmark & \xmark & 37.6 & 46.2 & 25.5 & 37.8 \\
FedBN & Supervised TENT & \cmark & \xmark & 40.1 & 47.9 & 29.2 & 39.0 \\
FedBN & FedPDA: FedBN & \cmark & \xmark & 39.6 & 49.9 & 31.2 & 40.2 \\
FedBN & FedPDA: PCGrad & \cmark & \xmark & 38.4 & 48.3 & 22.4 & 36.4 \\
FedBN & FedPDA: StarAlign & \cmark & \xmark & \textbf{44.4} & \textbf{53.6} & \textbf{33.5} & \textbf{43.8} \\ \midrule
None & FDA-KD3A & \xmark & \cmark & 41.4 & 47.4 & 31.8 & 40.2 \\ \bottomrule
\end{tabular}}
\end{table}

\subsection{Results and discussion}
\label{subsec:Results}

\textbf{Camelyon17 results - Table~\ref{tab:resultsCamelyon}:} \texttt{StarAlign} achieves the highest average accuracy over 5 hospitals among all methods, only outperformed in 1 out of 5 settings. In our FedPDA framework, replacing \texttt{StarAlign} with the prominent gradient alignment method \texttt{PCGrad}, degrades the performance. This shows the potential of our proposed gradient alignment method.\\
\textbf{Skin lesion diagnosis results - Table~\ref{tab:resultsSkinLesion}:} We observe that performance of all methods in this task is lower than in Camelyon17. This is a much more challenging task due to very high class-imbalance and strong domain shift\footnote{Our baselines achieve Accuracy 75-90\% on source domains of the skin lesion task, comparable to existing literature \cite{CASSIDY2022102305}, indicating they are well configured.}. It is extremely challenging to learn to predict the rarest minority classes under the influence of domain shift from very limited target data, which influences weighted-accuracy greatly. Accomplishing improvements in this challenging setting demonstrates promising capabilities of our method even in highly imbalanced datasets.

\texttt{StarAlign} consistently outperforms all compared methods. This includes PDA methods that cannot use source data (target \texttt{fine-tuning} and \texttt{Supervised TENT}) and the state-of-the-art FDA method \texttt{KD3A}, showing the potential of FedPDA. When FedPDA is performed with \texttt{StarAlign}, it outperforms FedPDA performed simply with a second round of \texttt{FedBN} training when the target node is connected to the FL system along with source nodes. \texttt{FedBN} can be viewed as Personalised \Gls{fl} method, as it learns one model per client via client-specific BN layers. Results demonstrate that learning one target-specific model with \texttt{StarAlign} yields better results. It also outperforms FedPDA with \texttt{PCGrad}, showing our gradient alignment method's effectiveness. This shows \texttt{StarAlign}'s potential to adapt a model to a distribution with high class imbalance and domain shift.

\section{Conclusion}

This work presents FedPDA, a framework unifying \Gls{fl} and \Gls{pda} without data exchange. FedPDA enables a pre-trained, deployed model to obtain information from source data via remote gradient communication to facilitate post-deployment adaptation. We introduce \texttt{StarAlign} which aligns gradients from source domains with target domain gradients, distilling useful information from source data without data exchange to improve adaptation. We evaluated the method on two multi-center imaging databases and showed that \texttt{StarAlign} surpasses previous methods and improves performance of deployed models on new domains.
\subsubsection{Acknowledgment:} Felix Wagner is supported by the EPSRC Centre for Doctoral Training in Health Data Science (EP/S02428X/1), by the Angela Krosik Award from the Anglo-Austrian Society, and by an Oxford-Reuben scholarship. Pramit Saha is supported in part by the UK EPSRC Programme Grant EP/T028572/1 (VisualAI) and a UK EPSRC Doctoral Training Partnership award.
The authors also acknowledge the use of the University of Oxford Advanced Research Computing (ARC) facility in carrying out this work (http://dx.doi.org/10.5281/zenodo.22558).

%
%
%
\bibliographystyle{splncs04}
\bibliography{references}
\end{document}